\documentclass[11pt,a4paper]{article}
\usepackage[hyperref]{acl2018}
\usepackage{times}
\usepackage{latexsym}
\usepackage{amsmath}
\usepackage[colorinlistoftodos]{todonotes}
\usepackage{multirow}

\usepackage{url}

\aclfinalcopy 

\title{Getting the subtext without the text: Scalable multimodal\\ sentiment classification from visual and acoustic modalities}

\author{Nathaniel Blanchard \\
  Dept. of Comp. Sci. and Eng. \\
  University of Notre Dame, USA \\
  {\tt nblancha@nd.edu} \\\And
  Daniel Moreira \\
  Dept. of Comp. Sci. and Eng. \\
  University of Notre Dame, USA \\
  {\tt dhenriq1@nd.edu} \\\AND
  Aparna Bharati \\
  Dept. of Comp. Sci. and Eng. \\
  University of Notre Dame, USA \\
  {\tt abharati@nd.edu} \\\And
  Walter J. Scheirer \\
  Dept. of Comp. Sci. and Eng. \\
  University of Notre Dame, USA \\
  {\tt walter.scheirer@nd.edu}
  }

\date{}

\begin{document}
\maketitle
\begin{abstract}
In the last decade, video blogs (vlogs) have become an extremely popular method through which people express sentiment. The ubiquitousness of these videos has increased the importance of multimodal fusion models, which incorporate video and audio features with traditional text features for automatic sentiment detection. Multimodal fusion offers a unique opportunity to build models that learn from the full depth of expression available to human viewers. In the detection of sentiment in these videos, acoustic and video features provide clarity to otherwise ambiguous transcripts. In this paper, we present a multimodal fusion model that exclusively uses high-level video and audio features to analyze spoken sentences for sentiment. We discard traditional transcription features in order to minimize human intervention and to maximize the deployability of our model on at-scale real-world data. We select high-level features for our model that have been successful in non-affect domains in order to test their generalizability in the sentiment detection domain. We train and test our model on the newly released CMU Multimodal Opinion Sentiment and Emotion Intensity (CMU-MOSEI) dataset, obtaining an $F_1$ score of 0.8049 on the validation set and an $F_1$ score of 0.6325 on the held-out challenge test set. 
\end{abstract}

\section{Introduction}

\begin{figure}
\centering
\includegraphics[scale=.365]{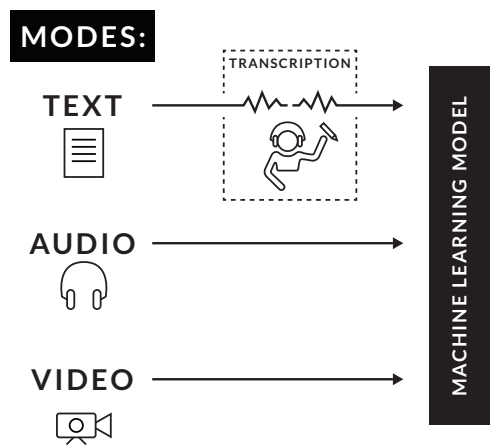}
\caption{A blindspot in multimodal sentiment analysis is the inclusion of human-transcriptions of spoken sentiment, which limits model applicability. We address this by using only prosodic and visual features for sentiment classification.}
\label{fig:teaser}
\end{figure}

Multimodal fusion models in the spoken-word domain incorporate features outside of text-based natural language processing (NLP) to increase model performance. These models benefit from the full scope of person--person interaction, which provides both context and clarification for speech that is ambiguous as text alone. The addition of multimodal data has been shown to increase model performance across a broad set of spoken-word fields, such as sarcasm \cite{joshi_automatic_2017}, question \cite{donnelly_words_2017} and sentiment \cite{zadeh_tensor_2017} detection. Each of these examples contains speech that can be difficult to infer from transcribed text---instead, the speaker's intent is clarified to listeners via intonations or expressions. It follows that machine learning models trained to include domain knowledge from these modalities would likewise be able to correctly interpret complex communication. 

Multimodal sentiment analysis (MSA) is one example of ambiguous speech that has been shown to benefit from additional modalities \cite{zadeh_tensor_2017,zadeh_memory_2018,chen_multimodal_2017,poria_context-dependent_2017,yu_end--end_2017}. MSA is the identification of the explicit or implicit attitude of a thought or sentence toward a situation or event. In recent years, the online community has been shown to frequently express sentiment orally in videos or recordings uploaded to sites like Youtube or Facebook. These spoken-word opinion pieces have been collected and annotated into large high-quality multimodal sentiment datasets \cite{zadeh_multimodal_2016, busso_iemocap:_2008, perez-rosas_utterance-level_2013, wollmer_youtube_2013, park_computational_2014}. 
Recently, the largest annotated sentiment dataset to date, CMU-MOSEI, was released \cite{cmumoseiacl2018}. This dataset contains over 23,500 spoken sentence videos, totaling 65 hours, 53 minutes, and 36 seconds. This large quantity of data comes from real-world expressions of sentiment, offering a unique opportunity to train and test model performance and generalization on a large dataset. Additionally, \citet{zadeh_multi-attention_2018} released a software development kit (SDK) for training and testing models on the CMU-MOSEI dataset, with future work focusing on addition of other multimodal datasets. These releases culminated in a challenge focused on human multimodal language with the opportunity to train a model and evaluate it on a held-out challenge test set. 

As is common in sentiment datasets, the MOSEI dataset includes features from human transcriptions of speech \cite{soleymani_survey_2017, poria_review_2017}. Ideally, models trained to annotate sentiment will operate on real-world data with as few barriers to deployment as possible in order to maximize efficiency and continuity. The use of human transcripts represents one of these barriers---it greatly limits the scalability of models in the real-world due to the time and cost in transcription and the inequality in quality between human and computer transcripts \cite{morbini_which_2013,blanchard_study_2015}.

The goal of this work is to build a model that broadly generalizes to unseen data using only scalable audio and visual features, reducing the need for transcription of human speech. In order to achieve this, we implement a model pipeline which has been successfully deployed in domains of sensitive and affectively impactful video analysis \cite{moreira_multimodal_2019}. From this pipeline, we select simple high-level video features and a generalized subset of audio features extracted using openSMILE \cite{eyben_opensmile:_2010}. We further test the generalizability of this pipeline by evaluating its applicability to the MSA domain. 

Additionally, this pipeline automatically extracts interpretable features that highlight model attention. These features can be easily mapped back to videos, as shown by \citet{moreira_pornography_2016}, which allows easy interpretation of model performance. Although recent work in MSA has begun exploring applicability of deep learning features, these models mostly achieve high performance numbers in specific scenarios but have poor generalizability and interpretability \cite{poria_multimodal_2018}.

In the next section, we examine related work on multimodal sentiment analysis. Section 3 explains the model pipeline and evaluation procedure. Section 4 presents our model results on the CMU-MOSEI validation set and the grand challenge held-out test set. Finally, in Section 5 we discuss our results, our model's limitations, and propose future work to improve our model. 
%
%
%

\begin{figure*}[t]
\centering
\includegraphics[scale=.65]{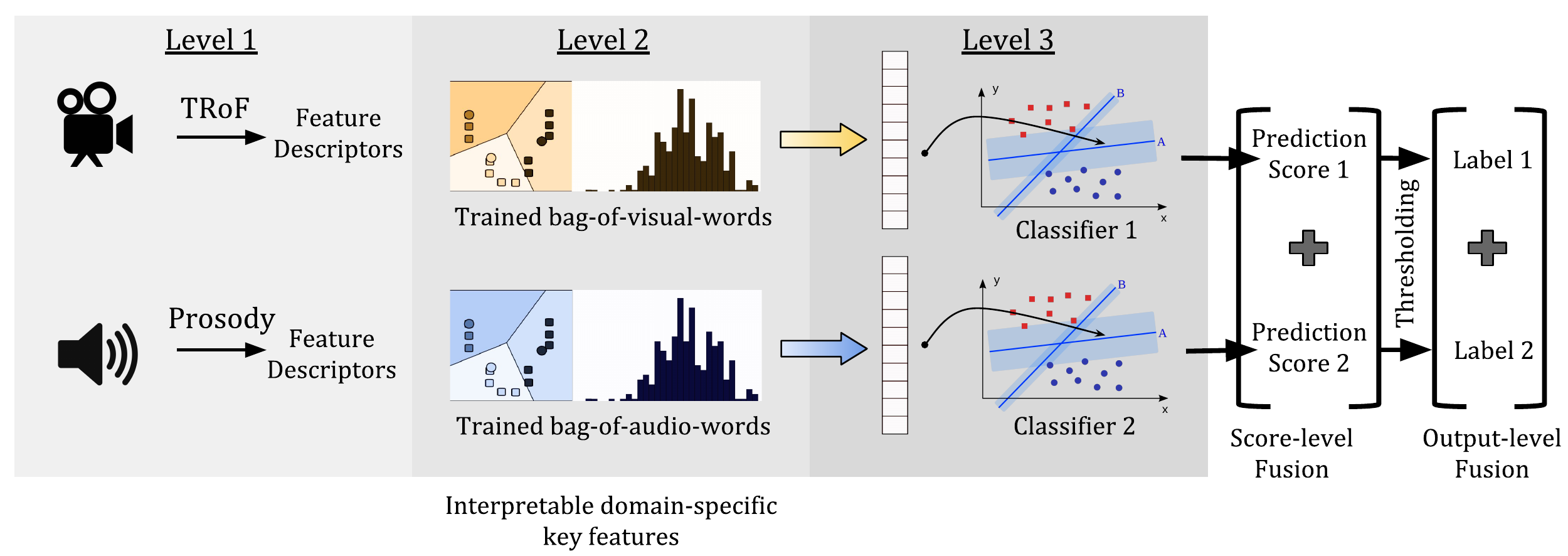}
\caption{Pipeline of our method with three stages: (i) low-level description of audio/visual stream, (ii) mid-level description of audio/visual stream using a trained bag-of-words model for each modality and (iii) training a classifier to predict class of the features. Fusion can be performed using the second stage features, the third stage score prediction or the thresholded score output labels.}
\label{fig:pipeline}
\end{figure*}
%
%
%
%

\section{Related Work}

Traditionally, sentiment analysis has been considered a natural language processing (NLP) problem, with data that largely consists of transcribed speech or written essays. The rise of YouTube and other video websites has facilitated an increase in multimodal forms of sentiment expression leading to the release of a number of high-quality video datasets annotated for sentiment \cite{zadeh_multimodal_2016, busso_iemocap:_2008, perez-rosas_utterance-level_2013, wollmer_youtube_2013, park_computational_2014}. These datasets have in turn led to an increased interest in multimodal fusion of video, audio, and text modalities for multimodal sentiment analysis (MSA), as summarized in recent surveys \cite{soleymani_survey_2017, poria_review_2017}. 

\subsection{Sentiment Analysis in the Wild}

A known issue with multimodal sentiment analysis (MSA) is the overemphasis on text features as opposed to visual or audio clues \cite{poria_review_2017}. In spoken sentiment, text restricts the applicability of the model in the wild due to human labor costs of transcription. However, given that a large majority of multimodal sentiment datasets include transcriptions, it is understandable that most researchers in this field have included these features in their models. Rather than minimizing this text-based work, our goal is instead to increase focus on audio and visual modalities as a key area for future MSA research. In this way, we are able to emphasize the real-world scalability of our model by excluding text features. Thus, we limit our review of previous work to recent applications of multimodal fusion of audio and visual features for MSA. 

Recent work in MSA using only audio and visual features is relatively sparse, despite the swath of such models in emotion detection \cite{poria_review_2017}. 
\citet{poria_deep_2015} extracted a multitude of frame-level video features and sentence-level audio features for multimodal fusion. They used feature selection to optimize classification of sentiment polarity (positive, negative, or neutral sentiment) and built an audio-visual model that achieved a validation accuracy of 83.69\%. Unfortunately, their study contains minimal focus on interpretability of generated and selected features. 

\citet{poria_multimodal_2018} recently published work that established baseline performance on MSA across a range of models and datasets. Their findings confirmed that multimodal audio and visual models have lower performance than multimodal models that contain text. They also found that MSA model performance plummets, regardless of modality, on cross-dataset tests. Additionally, \citet{poria_multimodal_2018} presented a machine learning model using audio and video features. They extracted video features using 3D convolution filters, and selected relevant features with a max-pooling operation. Their audio-visual model was evaluated on a variety of datasets, achieving accuracies between 67.90\% and 78.80\%, depending on the dataset and on training with same-speaker inclusion or not. Additionally, they report their results for various modality fusion techniques, with scores ranging between 58.6\% and 65.3\%.  

The difference between our work and these related works is our method of extracting features from video and representing video segments, which will be detailed in the next section. Additionally, the related works use a large set of audio features provided by openSMILE \cite{eyben_opensmile:_2010}, while we employ only prosodic features for speech analysis, namely fundamental frequency, voicing probability, and loudness contours. 

\subsection{Model Inspiration and Success in Non-affect Domains}

Our approach follows the precedent set by \citet{moreira_multimodal_2019}, who developed a generalized multimodal framework that focuses on robust, hand-crafted features. Their architecture provides an efficient and temporally-aware technique for multimodal data processing and has been shown to generalize across different domains, achieving state-of-the-art performance in pornography and violence detection with no human intervention.
Inspired by the promising results of this work and interested in further domain applications for the framework, we extract similar features from the MOSEI dataset to build a scalable solution to sentiment analysis in our study.


\section{Methods}

Sentiment expression and interpretation comprise abstract and complex phenomena, whose translation to audio and visual characteristics is not straightforward.
To cope with such complexity in a computationally affordable way (i.e., small runtime and low-memory footprint) we employ a Bag-of-Features-based (BoF) solution to the multimodal sentiment analysis (MSA) domain. 
BoF models reduce raw data from a modality into a collection of key local features. This technique reduces the semantic gap between the low-level audio and visual data representation, and the high-level concept of sentiment.

Our model training pipeline is presented in Figure~\ref{fig:pipeline}. Broadly, the pipeline extracts key features from the sentiment sentence videos and computes a confidence score for each modality. Then, the pipeline performs multimodal score fusion, generating a sentiment prediction. Full details can be found in \citet{moreira_pornography_2016}. Aside from experimenting with different fusion techniques, we performed no hyperparameter tuning in order to test the model's domain adaptation. 

One limitation of this training architecture is that currently the feature extraction portion of the framework is only trainable on two class problems. Thus, we binarize sentiment into positive and negative classes. Ideally, this training process will be modified in the future -- for now, the ground truth scores are thresholded with values $>0$ being positive and $\leq0$ being negative.

The BoF-based feature processing portion of the pipeline is divided into three levels:

\textbf{Level 1: Low-level Feature Extraction.} At this stage we extract low-level features from raw data. In our case, the audio and video streams in the raw videos are first separated and segmented. Temporal Robust Features (TRoF)~\cite{moreira_pornography_2016} are then extracted from the video frames. TRoF works by considering Gaussian derivatives for both the spatially and temporally co-located pixels in a set of video frames. Thus, it isolates and captures important spatiotemporal portions for motion description. The pixels of these portions can then be sampled across space and time, prior to being described by regular Speeded-Up Robust Features (SURF)~\cite{bay_speeded-up_2008}. 

From the audio stream, we extract prosodic features using the sub-harmonic sampling algorithm provided by openSMILE~\cite{eyben_opensmile:_2010}. We limit our selection of audio features from openSMILE to correspond with essential features for speech analysis, namely fundamental frequency, voicing probability, and loudness contours of the audio waves. These features have been identified as important in related implementations of the pipeline \citet{moreira_multimodal_2019}. 

\textbf{Level 2: Mid-level Feature Extraction.} 
At this stage we employ a mid-level coding step that quantizes the low-level features according to codebooks. Codebooks are a modular way of representing important features that provide a coarser representation of the video content that is closer and aware of the binarized concept of sentiment. Separate codebooks are created for each modality. 
For each codebook, we estimate Gaussian Mixture Models (GMM) from one million low-level features, with half of of the features coming from negative-sentiment examples, and the other half coming from positive-sentiment examples. Both GMMs are comprised of 256 Gaussian distributions.
After quantization, using the codebook, a pooling step summarizes all of the the mid-level features into a single feature vector for each video segment. 

Interpretable features can be extracted from the pipeline using the learned codebook, as described by \citet{moreira_pornography_2016}. 

\textbf{Level 3: Confidence Generation.} 
Once we obtain the mid-level feature vector for each of our video sentences, a separate linear Support Vector Machine (SVM) classifier is trained for each data modality.
In order to optimize the SVM for classification accuracy, we perform a 5-fold cross validation and select the best $C$, a SVM hyperparameter, using a $log_2$ scale in the range [-3,15]. 
Confidence scores are generated using the distance of the samples from the boundary learned by the classifier during training. These scores are then normalized between 0 and 1. 

\subsection{Prediction Using Multimodal Fusion}
Once we obtain confidence scores for each video segment, we employ two late fusion techniques to predict the class of each of the segments. Our methods are inspired by the domain of Biometrics ~\cite{ross_information_2003}, 
which has a long history of employing multiple modalities in real-world applications to improve model performance. 

\begin{enumerate}
	\item \textbf{Score-level Fusion} The normalized scores for video frame classification and audio signal classification are averaged to obtain our final classification scores. To further extend our fusion-based approach, the weight of each of the two scores contributing to the mean is treated as a hyperparameter, $\theta$. For the validation results we weight both the scores equally and threshold the scores at 0.5 to obtain labels. For evaluation on the test set, we choose the relative weight parameter corresponding to the most accurate validation results. The objective function used to optimize the hyperparameter is defined as:
    \begin{equation}
    \label{eqn:classificationerror}
    \operatorname*{arg\,min}_\theta \dfrac{1}{N} \sum_{c=1}^{N} \dfrac{1}{n_c} \sum_{i=1}^{n_c} I (y_i \neq \hat{y_i})
    \end{equation}
    Here, $\hat{y_i}$ can be defined as:
    \begin{equation}
    \label{eqn:prediction}
    \hat{y_i} = th(\theta*vScore_i + (1-\theta)*aScore_i)
    \end{equation}
    
    Equation~\ref{eqn:classificationerror} denotes the average number of classification errors across all classes. $N$ represents the number of classes (in our case, 2) and $n_c$ corresponds to the number of samples belonging to class $c$. $y_i$ is the ground truth label and $\hat{y_i}$ is obtained by thresholding the weighted average score as presented in Equation~\ref{eqn:prediction}. $I(.)$ is an indicator function that takes values 1 when $y_i$ is equal to $\hat{y_i}$. $th(.)$ is the thresholding function that uses $(1-\theta)$ as the threshold corresponding to each value of $\theta$ in the equation. The optimized hyperparameter was chosen after testing with grid search in the range [0,1] with a step of 0.2. Here, $\hat{y_i}$ for $\theta = 0$ and $\theta = 1$ correspond to unimodal (either video or audio) classification labels.
    
\item \textbf{Output-level Fusion} This is a simple fusion technique applied through the method of thresholding all of the scores obtained from our classifiers. The thresholded scores are $\in \{-1,0,1\}$ and are applied upon uniform binning of the raw confidence scores. We added the thresholded scores for both our modalities and scaled them to a range of 0 to 1. This score was then able to act as the predicted score for a video to belong to a particular class.
\end{enumerate} 

\subsection{MOSEI}

For this work we trained, tested, and validated our model on the MOSEI dataset \cite{cmumoseiacl2018}. The dataset was composed of over 23,500 spoken sentence videos, totaling 65 hours, 53 minutes, and 36 seconds. The dataset had been segmented at the sentence level; the sentences had been transcribed, and audio, visual, and textual features had been generated and released as part a public \citet{zadeh_multi-attention_2018} software development kit (SDK). Additionally, raw videos were available for download. Each video had been human scored on two levels: sentiment, which ranges between [-3,3], and emotion, which had six different values. For the purpose of this work, we focused only on the sentiment scores. 

For our purposes we extracted features from the raw videos and used the SDK to obtain the dataset's training, testing, and validation sets. 

\begin{table*}[t] 
\renewcommand{\arraystretch}{1.5}
\caption{Performance of individual modality and multimodal fusion for sentiment analysis on the validation set of CMU-MOSEI. MAE is the Mean Absolute Error.}
\vspace{5pt}
\centering
\begin{tabular}{|l|c|c|c|c|} 
\hline
    Solution & Precision & Recall & F1-Score & MAE\\ 
    \hline
    Audio Prosodic + SVM & 0.7485 & 0.4831 & 0.5872 & 0.7919\\ 
    Video TRoF + SVM    & 0.7928 & 0.7198 & 0.7545 & 0.7811\\ 
    Score-level Fusion    & \textbf{0.8022}  & 0.5749 & 0.6698 & 0.7849\\ 
    Output-level Fusion  & 0.7729 & \textbf{0.8396} & \textbf{0.8049} & \textbf{0.7760}\\ 
    \hline
\end{tabular}
\end{table*}

\subsection{Evaluation Metrics}
Our model output presents predictions as binary positive or negative classes as well as a confidence metric for each video sentence.

We evaluated our model's performance on basic classification of sentiment using precision, recall and $F_1$-scores. We selected these metrics because they are known to report accurate performance representation on imbalanced classes. Since these metrics are defined for two-classes, we binarize the ground truth scores values by thresholding values $>0$ as positive and the remaining as negative. 

Although we trained the SVM classifier for binary predictions, the confidence scores obtained from the classifier for each sample are continuous and can be used to perform regression. Since sentiment scores in the dataset scale between [-3, 3], we scaled our confidence scores to match the expected distribution of sentiment using a linear transformation function. These were the predictions that we submitted to the ACL2018 Grand Challenge. We also performed a regression between the ground truth scores and scores obtained by our methods on the validation set, and reported the Mean Absolute Error (MAE) for these experiments alongside our classification results. 

\section{Results}

In this section we present our results on both the MOSEI validation set and the ACL2018 Grand Challenge MOSEI test set. In the validation set section we report the evaluation metrics we used to assess the performance of our model and in the test set section, we present the metrics used by the ACL 2018 Grand Challenge organizers. 

\subsection{Validation Set Results}

Using the metrics of evaluation described in Section. 3.3, we tested our proposed approach on the validation set of the CMU-MOSEI dataset. In general, our model's performance was comparable to related work with the best method achieving F1-score of 0.80. The classification and regression results are presented in Table 1. A finer analysis of correct and wrong classification is presented in Table 3. The video portion of our model performed well on the validation and our fusion techniques resulted in improved performance with respect to using unimodal models. However, the audio-only model performed relatively poorly, indicating that our model's major weakness was in the audio domain. We expand upon this weakness in section 5.1.


\subsection{Test Set Results}

The classification metrics reported by the organizers on the test set include average $F_1$-score and average class accuracies considering different numbers of sentiment classes. For regression, they report MAE and the correlation coefficient between ground truth and prediction scores. In the regression scenario, our submission method (Fusion 1) obtained a MAE of 0.91 on the test set and 0.78 on the validation set. The specific metrics and the values achieved by our method on the test set have been reported in Table 2. 


\begin{table} 
\renewcommand{\arraystretch}{1.5}
\caption{Performance of the proposed approach in terms of the metric of evaluation used in ACL2018 Human Multimodal Language Challenge} 
\vspace{10pt}
\centering
\begin{tabular}{|c|l|}
\hline
Metric & Value \\\hline
Mean Average Error (MAE) & 0.9108 \\\hline

Correlation Coefficient & 0.3051 \\\hline

Average Binary Accuracy & 0.6094 \\\hline

Average Weighted Binary Accuracy & 0.6108 \\\hline

Average F1 Score & 0.6325 \\\hline

Average 5-Class Accuracy & 0.3320 \\\hline

Average 7-Class Accuracy & 0.3296 \\\hline

\end{tabular}
\end{table}


We use a binary training technique and correspond the SVM confidence scores to sentiment intensities. However, these results suggest that continuity in our scores does not correspond well with quantized sentiment bins.
 

%
%
%

\begin{table} 
\renewcommand{\arraystretch}{1.5}
\caption{Confusion matrix of classification results from the methods on the validation set of CMU-MOSEI.} 

\vspace{10pt}
\centering
\renewcommand\arraystretch{1.2}
\begin{tabular}{|c|l|c|c|}
\hline
$\downarrow$ Predicted & Actual $\rightarrow$ & Positive & Negative \\ \hline
Positive & \begin{tabular}[c]{@{}l@{}}Audio\\ Video\\ Fusion 1\\ Fusion 2\end{tabular} & \begin{tabular}[c]{@{}c@{}}615\\ 884\\ 706\\ 1031\end{tabular} & \begin{tabular}[c]{@{}c@{}}613\\ 344\\ 522\\ 197\end{tabular} \\ \hline
Negative & \begin{tabular}[c]{@{}l@{}}Audio\\ Video\\ Fusion 1\\ Fusion 2\end{tabular} & \begin{tabular}[c]{@{}c@{}}181\\ 231\\ 174\\ 303\end{tabular} & \begin{tabular}[c]{@{}c@{}}290\\ 240\\ 297\\ 168\end{tabular} \\ \hline
\end{tabular}
\end{table}

\section{Limitations and Future Work}

In order to be deployed at scale in real-world scenarios, machine learning models should have minimal-to-no human intervention to becoming fully automated. We maximized the automation of our model by discarding human-transcription data, instead relying solely on audio and video features. While this is an important step, we identified three major limitations of our model that should be improved before it is deployed at-scale. First, the quality of our chosen integration of audio features resulted in a poor representation of sentiment. Second, our results show that SVM distance does not map well to sentiment intensity. Third, the CMU-MOSEI dataset pre-segments data into sentences and omits non-labeled segments. This makes it impossible to obtain a realistic representation of real-world data using only this dataset. By isolating and expanding on these obstacles and their effects on our model's performance, which we do below, we are able to come to noteworthy conclusions that can be incorporated into future work.

\subsection{Audio feature limitations}

Audio features for our model were selected based on comparison with related work \cite{moreira_multimodal_2019}. Unfortunately, our multimodal model received relatively little benefit from the audio modality when evaluated on the validation set. We suspect that a major reason for this failure is the relatively poor audio quality of the CMU-MOSEI dataset compared with the dataset used for the related work, which was comprised of production-level videos with Hollywood-level audio qualities. 

This is notable as an informative guide to the unforeseen limitations of the previous dataset that related work selected features on \cite{moreira_multimodal_2019}. Based on that dataset, we limited our model to three audio features. However, \citet{poria_deep_2015} built an audio model which used a large set of audio features (6,373 per video) to obtain a 74.49\% classification accuracy for positive, negative, and neutral sentiment. They found that feature selection, which typically improves accuracy, actually decreases audio model performance in the sentiment domain. This suggests that it is better to use as many audio features as possible when building MSA models. We briefly investigated adding more audio features by extracting 384 features from openSmile's emotion feature set \cite{schuller2009interspeech}. Unfortunately, this model only obtained an F1 of 0.51, compared to our model's 0.59. In future work we plan to experiment with audio features further in order to find what works best across domains. 

\subsection{SVM Distance Limitation}

As noted in the results section, the continuous scores generated for predictions using SVM are more granular than the ground truth sentiment scores. When the two are compared, the offset in the scores can lead to higher errors than if they were quantized in the same manner. Based on our observations, we would suggest usage of other techniques for extraction of sentiment intensity.

\subsection{Dataset Limitations}

The CMU-MOSEI dataset \cite{cmumoseiacl2018} used to train and test our model provides a large-scale breakdown of sentiment analysis. However, the dataset follows typical practices for multimodal sentiment datasets, which make it difficult to train a fully automatic model. We identify practices which would increase automation. First, the data is pre-segmented at the sentence level, resulting in no sentenceless data. For a model to be employed in the real-world, it needs to be aware of sentenceless data as well as imperfect sentence boundaries. For example, human often segment speech at the sentence or category level \cite{stolcke_dialogue_2000, zadeh_multimodal_2016}, however, machine learning algorithms have yet to perfect this practice. Previous work has found that NLP models are prone to complete failure when presented with excess words or information, even when those words are unrelated to the task \cite{jia_adversarial_2017}. Ideally, models in the real-world will be robust to such noise. 

Second, our model does not use human transcription in order to avoid limitations in real-world applicability. However, text is a modality that improves MSA. Rather than releasing text transcriptions for model building, we propose future datasets release automatic speech recognition transcriptions. This would further model automation by incorporating scalable transcription practices, as is becoming more common in other domains \cite{blanchard_semi-automatic_2016}. Additionally, recent work suggests the gap between human transcription and ASR will soon be negated by advances in the speech recognition domain \cite{stolcke_comparing_2017}, furthering the argument that human transcription is no longer necessary for building models. 

By including the full range of data and switching from human to ASR transcription, we believe that sentiment models can be trained, evaluated, and employed at-scale on real-world data. Work on automating multimodal sentiment analysis should focus on model performance using tractable methods of data collection; as exemplified by other domains intended to work with real-world data \cite{ram_conversational_2018, yan_multi-clue_2016}, with human level transcriptions of data reported as a comparison metric. 

\section{Conclusion}

We conclude our study with the presentation of the results of a generalized model for multimodal sentiment analysis using only visual and audio modalities. 
In this work, we completed two significant goals: first, we trained and evaluated a MSA model at scale with minimal human intervention. Second, we tested the cross-domain generalizability of a model framework that has shown great success in other multimodal domains. 
Although multimodal sentiment analysis has traditionally been characterized as a natural language processing field driven by human transcription, we believe that our results show the tractability of models built without human-in-the-loop. We advise researchers to ensure that their future work makes an effort to limit transcript-based datasets by employing automatic speech transcription. By doing this, they will be able to further minimize human interaction and allow their models to approach full automation. 
This work is one component of a broader effort in the MSA community to expand MSA to process real-world data at scale. Despite the limitations of our model, we believe that our work creates substantial groundwork for further investigation of video- and audio-based models. 

\bibliographystyle{acl_natbib}
\bibliography{Zotero.bib}

\end{document}